\documentclass[10pt,twocolumn,letterpaper]{article}

\usepackage[pagenumbers]{cvpr} % To force page numbers, e.g. for an arXiv version

\usepackage[dvipsnames]{xcolor}

\definecolor{cvprblue}{rgb}{0.21,0.49,0.74}
\usepackage[pagebackref,breaklinks,colorlinks,citecolor=cvprblue]{hyperref}
\usepackage{mathtools}
\usepackage{amsmath,bm}
\def\paperID{*****} % *** Enter the Paper ID here
\def\confName{CVPR}
\def\confYear{2024}

\title{Model-Agnostic Human Preference Inversion in Diffusion Models}

\author{Jeeyung Kim, Ze Wang, Qiang Qiu\\
Purdue University\\
{\tt\small \{jkim17, wang5026, qqiu\}@purdue.edu}
}

\begin{document}
\maketitle
\begin{abstract}
Efficient text-to-image generation remains a challenging task due to the high computational costs associated with the multi-step sampling in diffusion models. 
Although distillation of pre-trained diffusion models has been successful in reducing sampling steps, low-step image generation often falls short in terms of quality.  In this study, we propose a novel sampling design to achieve high-quality one-step image generation aligning with human preferences, particularly focusing on exploring the impact of the prior noise distribution.
Our approach, \textit{\textbf{P}rompt \textbf{A}daptive \textbf{H}uman Preference \textbf{I}nversion (\textbf{PAHI})}, optimizes the noise distributions for each prompt based on human preferences without the need for fine-tuning diffusion models. Our experiments showcase that the tailored noise distributions significantly improve image quality with only a marginal increase in computational cost. Our findings underscore the importance of noise optimization and pave the way for efficient and high-quality text-to-image synthesis.
\end{abstract}    
\section{Introduction}
\label{sec:intro}
Recent advances in diffusion models (DMs) have unlocked unprecedented capabilities in text-to-image generation~\citep{rombach2022high}. Despite their promise, the widespread adoption of DMs in practical applications may be hindered by the high inference costs associated with the multi-step sampling.
To tackle this challenge, considerable efforts have been focused on reducing the number of sampling steps while maintaining image quality. In particular, distilling pre-trained DMs using adversarial approaches \citep{sauer2023adversarial, yin2023one} and consistency regularization on ordinary differential equation (ODE) trajectories \citep{song2023consistency, luo2023latent} emerged as effective strategies. These methods enable high-fidelity text-to-image generation in a few steps. 
However, images produced in fewer steps often exhibit inferior quality compared to those generated with more steps~\citep{sauer2023adversarial}.

Prior studies~\citep{karras2022elucidating, song2020denoising, lu2022dpm} modify the sampling process to improve the quality of image generation without altering the training procedure. They propose methods to enhance the multi-step sampling in DMs, focusing on modifications to solve the probability flow (PF), such as adjustments to noise schedules~\citep{karras2022elucidating, lu2022dpm} and control of stochasticity levels~\citep{song2020denoising}.
Nevertheless, for distilled models with extremely limited sampling steps, alternative sampling designs—orthogonal to direct manipulation of the PF solver—can be essential to further advance image quality, particularly considering that the characteristics of DMs can diminish in distilled models.
One potential approach involves developing a sampling process combined with prompt optimization. \citep{hao2024optimizing, manas2024improving} introduce a prompt optimization framework that tailors user input to model-preferred prompts.

Another promising approach is optimizing noise prior, an additional input alongside prompts in DMs.
Notably, the prior noise distribution, typically fixed as a standard Gaussian, has been overlooked in optimization to enhance sampling quality.
Especially when the sampling process in DMs is deterministic, the prior noise directly shapes and influences the resulting images, underscoring the potential for adjusting the noise distribution for superior images.
Therefore, our study aims to discover a noise distribution that surpasses the standard Gaussian in generating higher-quality images.
We focus on one-step generation because it clearly reflects the impact of prior noise, unlike multi-step generation where the effect of prior noise can be gradually attenuated during sampling.

To discover the (sub-)optimal noise distribution, we employ a DM as an image generator and a scoring model~\citep{kirstain2024pick} as an evaluator to assess the generated image quality.
This scoring model was trained on human preferences regarding image-prompt pairs and plays a pivotal role in providing feedback on the alignment of generated images with human preferences, emulating a \textit{Human-in-the-Loop} framework.
Inspired by optimization-based \textit{inversion} techniques~\citep{abdal2019image2stylegan, zhu2020improved} commonly used in Generative Adversarial Networks (GANs)~\citep{goodfellow2014generative}, we directly optimize the parameters of noise distribution based on human preference scores, leading to enhancements in image quality while keeping DMs intact.

Taking our investigation a step further, we show that (sub-)optimal noise distributions can vary depending on the text prompts provided. 
We additionally introduce a crucial component: a noise predicting model. The light-weight model, comprising a pre-trained text encoder and shallow layers, processes text prompts and predicts parameters of a Gaussian distribution. This distribution is then employed as a prior noise for the image generator. 
We train the noise predicting model to produce noises that help DM generate images with high scores.
We demonstrate that the noise Gaussian distribution with parameters tailored to specific text prompts generates superior images in one step, aligning with human preference.

Our method involves inverting human preferred images back into the noise space, termed as \textit{\textbf{P}rompt \textbf{A}daptive \textbf{H}uman Preference \textbf{I}nversion (\textbf{PAHI})}. 
\textit{PAHI} serves as a \textit{model-agnostic} image enhancement approach by adjusting the noise distribution using a \textit{lightweight noise-predicting model}. 
Notably, our method is a general framework applicable in multi-step deterministic sampling scenarios, extending its potential impact beyond one-step sampling.
To the best of our knowledge, we are the first to investigate the impact of noise optimization on text-to-image synthesis. 
By bridging the efficiency of low-step generation with enhanced quality, we unlock the potential of diffusion models for real-world applications.

\section{Related Work}
\textbf{Distilling diffusion models for low-step sampling.} Numerous studies \citep{meng2023distillation, song2023consistency, luo2023latent, liu2023instaflow, sauer2023adversarial, yin2023one, salimans2022progressive, yin2023one} focused on accelerating sampling process of DMs.
Among them, \citet{song2023consistency} introduces \textit{consistency models}, where points on the same trajectory of ODE map to the same initial point, enabling low-step generation. However, this method has not yet been applied to text-to-image synthesis. In the realm of text-to-image generation, \citet{luo2023latent} applies the consistency model to latent space and facilitates text-to-image generation by distilling Stable Diffusion~\citep{rombach2022high}. Another notable approach to regulating trajectories of ODEs is \citet{liu2023instaflow} introducing the concept of Rectified Flow, aimed at straightening the trajectories of ODEs. On the other hand, \citet{sauer2023adversarial} employs adversarial loss and score distillation sampling loss~\citep{poole2022dreamfusion} to generate images, showcasing superior performance compared to other distillation methods. However, low-step (1 or 2) generation often display inferior quality compared to multiple-step image generation.

\noindent\textbf{Prompt adaptive sampling design.}
\citet{hao2024optimizing}  introduce prompt adaptation, adjusting user input to match model-preferred prompts without DM fine-tuning. During sampling, text inputs undergo prompt adaptation processing before being fed into diffusion models.
\citet{zhang2023adadiff} emphasize the significance of employing diverse sampling steps relying on prompts and propose a framework of sampling with instance-specific steps to reduce sampling cost without compromising image quality.
However, none of these studies explore the potential of optimizing noise distribution contingent upon prompts.

\noindent\textbf{Fine-tuning diffusion models on human feedback.}
\citep{wallace2023diffusion, fan2024reinforcement, deng2024prdp} align DMs with human preferences by directly optimizing them based on human comparison data using reinforcement learning. However, such approaches differ from our model-agnostic approach, as it requires fine-tuning DMs, which can be resource-intensive. 

\noindent\textbf{Inversion.}
In prior studies, \textit{inversion} has been crucial for image manipulation, aiming to find a latent representation corresponding to a given image. In the GAN literature, \citep{abdal2019image2stylegan, zhu2020improved, gu2020image, roich2022pivotal}  use optimization-based techniques, directly optimizing latent vectors, while \citep{richardson2021encoding, zhu2020domain, pidhorskyi2020adversarial} employ trained encoders to map images to their latent representations.
In DMs, \textit{inversion} is adapted to accommodate their properties like stochastic, multi-step sampling and conditioned generation.
\textit{Textual Inversion}~\citep{gal2022image} represents user-provided concepts as pseudo-words in the text embedding space for versatile editing.
\citep{song2020denoising} proposes a deterministic sampling which enables inversion in a closed-form manner, facilitating image manipulation in DMs~\citep{song2020score, ramesh2022hierarchical, mokady2023null, kim2022diffusionclip,zhang2024real}.

In contrast to prior works, our study does not use inversion technique for the purpose of image manipulation. Instead, we invert human preferences into noise space, where samples from this space lead to improved image quality.
\section{Method}
\label{sec:formatting}

\subsection{Preliminaries}
In this study, we use the distilled diffusion model (\textit{Stable Diffusion 2.1} backbone~\citep{rombach2022high}) trained with adversarial loss~\citep{sauer2023adversarial}, which exhibits superior performance in low-step image generation. 
Similar to general diffusion models, the generative process of the distilled model progressively denoises a noisy observation starting from the standard Gaussian where $p(\bm{x}_T) = \mathcal{N}(\bm{x}_T; \bm{0},\bm{I})$ and $T$ represents the total number of timesteps during the training of the diffusion model.
The denoised observation is predicted with the distilled diffusion model ($\epsilon_{\theta}$)~\citep{ho2020denoising} as followed:
\begin{equation}
    \text{SD}_\theta(\bm{x}_{\tau_l}, c) := \frac{\bm{x}_{\tau_l} - \sqrt{1-\alpha_{\tau_l}} \cdot \epsilon_{\theta} (\bm{x}_{\tau_l}, c)}{\sqrt{\alpha_{\tau_l}}},
\end{equation}
where $\text{SD}_\theta(\cdot, \cdot)$ represents a denoised image, $l \in \{1, \cdots, L\}$, $\tau_L=T$ and $L$ is set to values less than 4 in \citep{sauer2023adversarial}. $c$ denotes conditions and $\alpha_{\tau_l}$ denotes a variance schedule at $\tau_l$.
Note that our study primarily concentrates on one-step generation ($L=1$).

On the other hand, \citet{kirstain2024pick} create an open dataset of text prompts and real users’ preferences over generated images. Based on this dataset, they train a CLIP-based \citep{radford2021learning} scoring model, named PickScore, which shows superior alignment with human preferences on generated images.
We use PickScore as our scoring model in the following sections. 
Note that alternative models can also be employed as a scoring model.

\subsection{The Proposed Method (PAHI)}
In this section, we introduce our proposed method which enhances image quality while preserving intact diffusion models.
We first propose a human preference inversion method in which a single noise distribution is optimized for all prompts. Following this, we propose \textit{\textbf{P}rompt \textbf{A}daptive \textbf{H}uman Preference \textbf{I}nversion (\textbf{PAHI})} that predicts customized noise distributions for individual prompts.

\noindent \textbf{Optimizing noise distribution across all prompts.}
We employ a distilled diffusion model for image generation and a scoring model to evaluate the generated images.
The output of scoring model reflects predicted human preference for the generated images.
The score of the generated image given the text prompt $c_i$ is defined as followed:
\begin{equation}
\label{eq:pipeline}
    s(\bm{x}_T^m, c_i)= \text{SC}_\phi (\text{SD}_\theta(\bm{x}_T^m, c_i), c_i), 
\end{equation}
where $\text{SC}_\phi$ is a scoring model, $\bm{x}_T^m$ denotes the $m$-th sample from $p(\bm{x}_T)$, $c_i \in \{c_1, c_2, \dots, c_n\}$ and $n$ denotes the number training text prompts. We interchangeably use $s(\bm{x}_T^m, c_i)$ and $s(\bm{x}_T, c_i)$.

We posit the existence of a potentially superior Gaussian distribution ($p(\bm{x}_T')$) compared to the standard Gaussian, which satisfies
\begin{equation}
    p(\bm{x}_T') = \mathcal{N}(\bm{x}_T; \bm{\mu}, \text{diag}(\bm{\sigma})), \; s(\bm{x}_T',c_i) > s(\bm{x}_T,c_i).
\end{equation}
Thus, we aim to optimize $\bm{\mu}$ and $\bm{\sigma}$ of the prior noise ($X_T'$) to maximize scores (align with human preference),  where $\bm{\mu}, \bm{\sigma} \in \mathcal{R}^{4  k^2}$ and $k$ denotes the size of latent variable in the pre-trained latent diffusion model~\citep{rombach2022high}.

We find a superior Gaussian distribution ($p(\bm{x}_T')$) by minimizing the objective function defined as followed:
\begin{align}
\label{eq:objective}
   \mathcal{L} = - \sum^n (0\cdot \log &f(\bm{\mu}, \bm{\sigma}) + 1\cdot \log f'(\bm{\mu}, \bm{\sigma})),\\
   \bm{\mu^*}, \bm{\sigma^*} &= \underset{\bm{\mu}, \bm{\sigma}}{\mathrm{argmin}} \; \mathcal{L},
\end{align}
where $f(\bm{\mu}, \bm{\sigma}) = \frac{e^{s}}{e^{s}+e^{s'}} $ and  $f'(\bm{\mu}, \bm{\sigma}) = \frac{e^{s'}}{e^{s}+e^{s'}}$.  
We use $s'$ instead of $s(\bm{x}_T', c_i)$ and $s$ instead of $s(\bm{x}_T, c_i)$ for brevity.

To optimize parameters of Gaussian distribution, we use the reparameterization trick~\citep{kingma2013auto} as followed:
\begin{align}
\bm{x}_T'^j= \bm{\epsilon}' \odot \bm{\sigma}^2 + \bm{\mu}, \; \text{where} \;  \bm{\epsilon}' \sim N(\textbf{0}, \textbf{I}),
\end{align}
where $\bm{x}_T'^j$ denotes $j$-th sample from $p(\bm{x}'_T)$.
The sample is used as input to $\text{SD}_\theta$ alongside the text prompt $c_i$.
We use the identified $\bm{\mu^*}$ and $\bm{\sigma^*}$ for all prompts during inference. 

\noindent \textbf{Prompt-adaptive noise distribution.}
We take one step further to tailor $\bm{\mu}(c_i)$ and $\bm{\sigma}(c_i)$ for an individual prompt $c_i$.
Built upon the previous framework, we further construct a conditional noise prediction model consisting of a pre-trained text encoder ($E$) and respective 2 MLPs ($g_{\psi}$) for predicting $\bm{\mu}(c_i)$ and $\bm{\sigma}(c_i)$.
The $g_{\psi}$ takes text embedding as input and outputs predictions of parameters of a Gaussian distribution.

The entire procedure is as followed:
\begin{align}
    g_{\psi}(E(c_i)) = &(\bm{\mu}(c_i), \bm{\sigma}(c_i)),\\
    \bm{x}_T'^j = \bm{\epsilon}' \odot \bm{\sigma}(c_i)^2 + &\bm{\mu}(c_i),  \; \text{where} \;  \bm{\epsilon}' \sim N(\textbf{0}, \textbf{I}),\\
    s(\bm{x}'_T, c_i)= \text{SC}_\phi & (\text{SD}_\theta(\bm{x}_T'^j, c_i), c_i),
\end{align}
where we use the text encoder of the diffusion model ($\epsilon_\theta$) for $E(\cdot)$ not to impose additional computation.
We find $\psi^*$ that minimizes objective function defined in \cref{eq:objective}.
During inference, we initially identify the optimal $\bm{\mu}^*(c_i)$ and $\bm{\sigma}^*(c_i)$ for prompt $c_i$ using the noise predicting model and then employ the Gaussian as a prior noise for the distilled diffusion models to generate images.

When $\psi$ are initialized randomly, $\bm{\mu}(c_i)$ and $\bm{\sigma}(c_i)$ may significantly deviate from their appropriate values, $\bm{0} \;\text{and} \;\bm{I}$. To stabilize the initial training phase, we opt to pre-train $g_{\psi}$ to produce parameters close to a standard Gaussian distribution while retaining the input text embedding information. This is achieved by minimizing the Kullback-Leibler (KL) divergence between the standard Gaussian and the Gaussian with the predicted parameters as well as a reconstruction loss of the text embeddings. We reconstruct text embedding using a decoder ($h_\omega$), 2-layer MLP, taking as input a sample of $\bm{x}'_T$ from $g_\psi$ and producing reconstructed text embeddings as output.

The loss function used for pretraining is structured as followed:
\begin{align}
   \mathcal{L'} = \sum_i^n \text{KL}(N(\bm{\mu}(c_i), \bm{\sigma}(c_i)), \; &N(\textbf{0}, \textbf{I})) \nonumber \\
   + \; \text{MSE}(E(c_i)&, h_{\omega}(g_{\psi}(E(c_i))).
\end{align}
We optimize $\psi$ and $\omega$ by minimizing the loss function. Subsequently, the updated parameters $\psi$ serve as the initialization of our noise prediction model for further training.

\section{Experiment}
We validate our framework (\textbf{\textit{PAHI}}) by demonstrating that the images generated by our method exhibit enhanced quality, based on one-step generation.

\noindent\textbf{Experiment setups.}
We conduct a comparison between images generated by our method and ones generated using the standard Gaussian distribution as a prior. 
We employ the dataset proposed by \citep{kirstain2024pick}, which consists of 35,000 distinct prompts. We randomly select 500 prompts for validation and another 500 prompts for the test set, while the remaining prompts are used for training. The scoring model used for training is PickScore~\citep{kirstain2024pick}. 
During evaluation, we employ both the PickScore and ImageReward scoring models~\citep{xu2024imagereward} to assess image quality and determine whether images optimized with PickScore align effectively with the criteria of ImageReward.
As an evaluation metric, we employ the win rate, determined by comparing the scores of two images generated from the same prompt: one using a predicted noise distribution and the other using the standard Gaussian distribution.
We denote our prompt-adaptive inversion method as {\textit{PAHI}}, while using a single inversion across all prompts is referred to as {\textit{HI}}.

\noindent\textbf{Implementation details.}
Our implementation is built upon the \textit{Huggingface Diffuser} framework\footnote{https://huggingface.co/docs/diffusers/en/index}.
We integrate \textit{ADD-M}~\citep{sauer2023adversarial}, where we reduce the computation and memory costs by replacing the VAE with a tiny autoencoder\footnote{https://github.com/madebyollin/taesd}. 
For \textit{PAHI}, we use the text encoder of the employed diffusion model as $E(\cdot)$ as it removes the need for additional computation. 
We generate images with a size of 512$^2$.
We configure the batch size to 72 and implement a learning rate warm-up for 10,000 steps, gradually increasing from $1 \times 10^{-5}$. We employ the Adam optimizer.
We use early stopping based on the average evaluation loss from Pickscore and Imagereward scoring models. 
If this loss does not drop for five consecutive evaluations, conducted every 1000 steps, training stops.
The inference time is measured using a Nvidia RTX 3090 GPU.

\subsection{Results}
\begin{table}
\vspace{-5mm}
    \centering
    \caption{The win rate against images generated from the standard Gaussian using different scoring models. The numbers represent the average (std) win rates of 5 runs with different seeds.}
    \begin{tabular}{c|c|c}
    \toprule
         & PickScore~\citep{kirstain2024pick}   & ImageReward~\citep{xu2024imagereward}    \\
         \midrule
        \textit{PAHI}& \textbf{94.0\%} (0.2) & \textbf{75.5\%}  (2.1) \\
        \textit{HI}&  64.7\% (1.6)  & 64.1\% (1.8)  \\
        % CLIP & &  \\
        \bottomrule
    \end{tabular}
    \label{tab:winrate}
\end{table}

\begin{table}
\vspace{-2mm}
\caption{Comparison of inference times between low step generation and our method: The numerical values represent the average time (scores) taken for an image sampling across 500 text prompts. We use batch size of 1.}
    \centering
    \begin{tabular}{c|c|c}
    \toprule
         &  Time ($s$) ($\downarrow$) &  Scores (PickScore) ($\uparrow$) \\
         \midrule
        \textit{PAHI}  & 0.067  &  \textbf{0.228} \\
        one step~\citep{sauer2023adversarial} & \textbf{0.062} &  0.212 \\
        two steps~\citep{sauer2023adversarial} &0.088& 0.212\\
        \bottomrule
    \end{tabular}
    \label{tab:efficiency_results}
\vspace{-3mm}
\end{table}

\noindent\textbf{Human preference scores comparison.} 
Our approach showcases superior performance compared to the baseline, as illustrated in Table \ref{tab:winrate}.
\textit{PAHI} significantly outperforms a standard Gaussian, with a remarkable win rate of 98.4\%.
\textit{HI} also shows effectiveness, achieving a win rate of around 65\%. However, its improvement does not match the efficacy of \textit{PAHI}.
Interestingly, even though trained with PickScore as the scoring model, \textit{PAHI} (\textit{HI}) excels in achieving higher ImageReward scores as well, boasting a win rate of 70.7\% (64.1\%).
These results demonstrate the effectiveness of optimizing noise distribution in enhancing image quality. Moreover, it underscores the importance of tailoring the noise distribution based on each prompt specifically.

\begin{figure}
\vspace{-5mm}
\centering
\begin{subfigure}{0.42\textwidth}
  \centering
  \includegraphics[width=.4\linewidth]{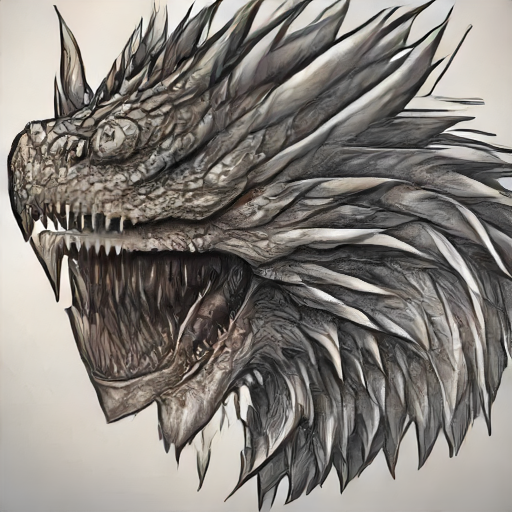}
  \hspace{6mm}
  \includegraphics[width=.4\linewidth]{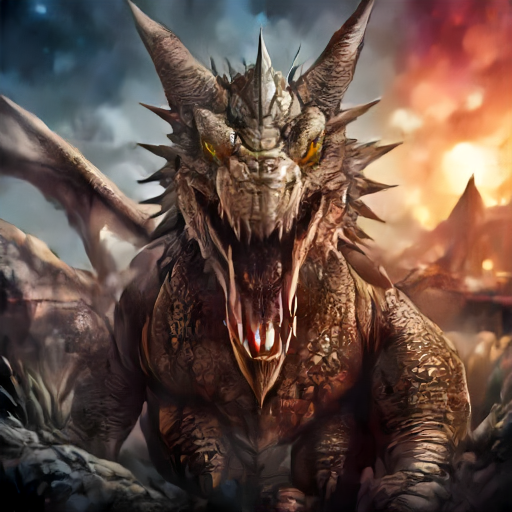}
  \caption{\bf{Head shot of a dragon, digital art style}}
  \label{fig:sfig1}
\end{subfigure}\\%
% \vskip\baselineskip
\begin{subfigure}{0.42\textwidth}
  \centering
  \includegraphics[width=.4\linewidth]{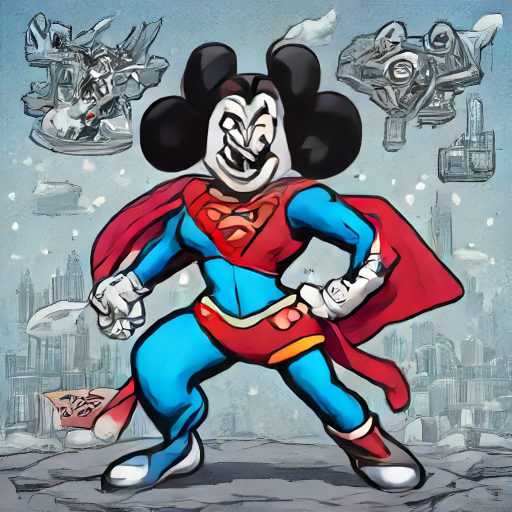}
 \hspace{6mm}
  \includegraphics[width=.4\linewidth]{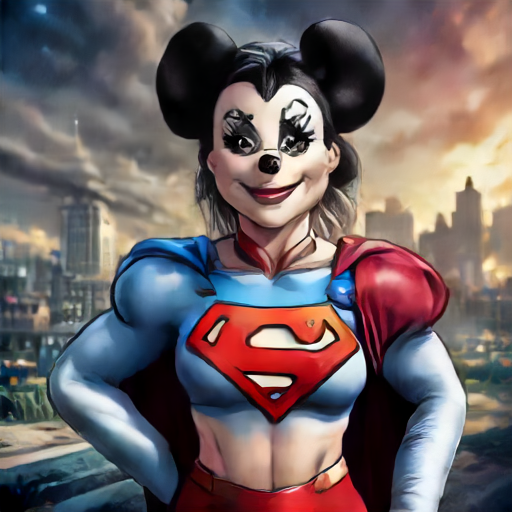}
  \caption{\bf{Minnie Mouse in a superman outfit bodybuilding, book illustration}}
  \label{fig:sfig2}
\end{subfigure}
\begin{subfigure}{0.42\textwidth}
  \centering
  \includegraphics[width=.4\linewidth]{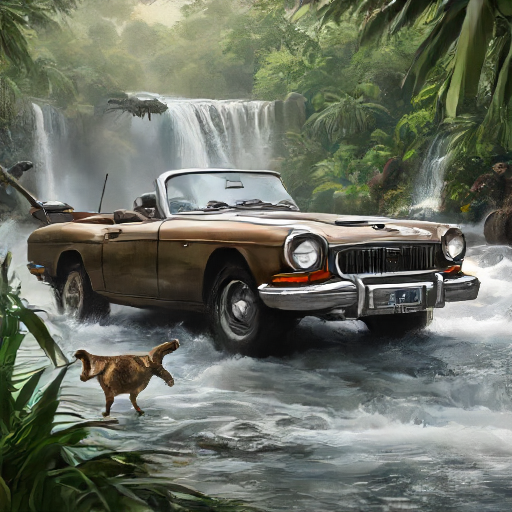}
  \hspace{6mm}
  \includegraphics[width=.4\linewidth]{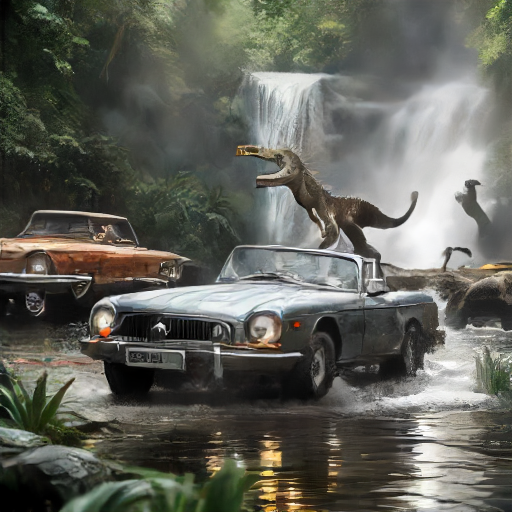}
  \caption{\bf{a velociraptor and an MGb in the jungle river, waterfall mist, Chrome Detailing}}
  \label{fig:sfig3}
\end{subfigure}
\caption{The prompts generated by users and the corresponding images sampled in one-step from the standard Gaussian (\textbf{left}) and the predicted noise distributions (\textbf{right}).}
\vspace{-4mm}
\label{fig:images}
\end{figure}

\noindent\textbf{Sampling computation cost.} 
We investigate the trade-off between image generation quality and sampling cost during inference. Typically, higher image quality is achieved with more sampling steps, which incur additional computation costs. 
We compare our approach with the standard Gaussian, employing 1 and 2 sampling steps~\citep{sauer2023adversarial}.
We present the time required for sampling to assess computational efficiency as well as the resulting scores to identify their quality, as depicted in Table \ref{tab:efficiency_results}. 
Despite no significant difference in scores between one-step and two-step generation,  two-step generation incurs additional time for sampling. 
In contrast, our approach demonstrates remarkable efficiency, requiring only a minimal increase in inference time compared to one-step generation (+0.005s), while showcasing higher scores aligned with human preferences.
When it comes to the number of parameters, \textit{PAHI} adds 5 million more parameters (two MLP layers) in total, which are negligible compared to the 983 million parameters of \textit{Stable Diffusion 2.1}~\citep{sauer2023adversarial}.
Note that we utilize the text encoder from the employed diffusion model, thus eliminating the necessity to introduce additional parameters to encode prompts.

\noindent\textbf{Generated images.} 
Figure \ref{fig:images} illustrates the generated images from our noise prediction and a standard Gaussian.
We underscore the images from the predicted noise (right) demonstrate improved image quality, aligning better with human preference.

\section{Conclusion}
Our study investigated improving the quality of text-to-image one-step generation.
We proposed a light-weight noise predicting model, optimizing noise distributions based on human preferences without fine-tuning diffusion models. 
Our experiments exhibited the tailored noise distributions improve image quality with marginal rise in computing cost. 
We highlight the efficacy of noise optimization, promising efficient and high-quality text-to-image synthesis.
{
    \small
    \bibliographystyle{ieeenat_fullname}
    \bibliography{main}
}


\end{document}